\documentclass[twoside]{article}

\usepackage{PRIMEarxiv}

\usepackage[utf8]{inputenc} 
\usepackage[T1]{fontenc}    
\usepackage{hyperref}       
\usepackage{url}            
\usepackage{booktabs}       
\usepackage{amsfonts}       
\usepackage{nicefrac}       
\usepackage{microtype}      
\usepackage{lipsum}
\usepackage{fancyhdr}       
\usepackage{graphicx}       
\graphicspath{{media/}}     
\usepackage{authblk}
\usepackage{amsmath}
\usepackage{multirow}

\title{Progressive Prompt-Guided Cross-Modal Reasoning for Referring Image Segmentation
}

\author[1]{Jiachen Li}
\author[2]{Hongyun Wang}
\author[1]{Jinyu Xu}
\author[3]{Wenbo Jiang}
\author[4]{Yanchun Ma}
\author[1]{Yongjian Liu}
\author[1]{Qing Xie}
\author[1]{Bolong Zheng}

\affil[1]{School of Computer Science and Artificial Intelligence, Wuhan University of Technology, Wuhan, China}
\affil[2]{School of Artificial Intelligence and Automation, Huazhong University of Science and Technology, Wuhan, China}
\affil[3]{University of Electronic Science and Technology of China, Chengdu, China}
\affil[4]{Wuhan Vocational College of Software and Engineering, Wuhan, China}

\begin{document}
\maketitle

\begin{abstract}
Referring image segmentation aims to localize and segment a target object in an image based on a free-form referring expression.
The core challenge lies in effectively bridging linguistic descriptions with object-level visual representations, especially when referring expressions involve detailed attributes and complex inter-object relationships.
Existing methods either rely on cross-modal alignment or employ Semantic Segmentation Prompts, but they often lack explicit reasoning mechanisms for grounding language descriptions to target regions in the image.
To address these limitations, we propose PPCR, a \textbf{P}rogressive \textbf{P}rompt-guided \textbf{C}ross-modal \textbf{R}easoning framework for referring image segmentation.
PPCR explicitly structures the reasoning process as a \textbf{Semantic Understanding-Spatial Grounding-Instance Segmentation} pipeline.
Specifically, PPCR first employs multimodal large language models (MLLMs) to generate Semantic Segmentation Prompt that capture key semantic cues of the target object.
Based on this semantic context, Spatial Segmentation Prompt are further generated to reason about object location and spatial extent, enabling a progressive transition from semantic understanding to spatial grounding.
The Semantic and Spatial Segmentation prompts are then jointly integrated into the segmentation module to guide accurate target localization and segmentation.
Extensive experiments on standard referring image segmentation benchmarks demonstrate that PPCR consistently outperforms existing methods. The code will be publicly released to facilitate reproducibility.
\end{abstract}

\keywords{Referring Image Segmentation \and Cross-modal Reasoning \and Prompt Learning \and Multimodal Large Language Models}

\section{Introduction}
\label{sec:intro}

Referring image segmentation~(RIS) aims to localize and segment the visual region in an image that corresponds to a referring expression, serving as a fundamental task in both visual grounding~\cite{Xiao_2025} and image segmentation~\cite{10613466}.
Free-form referring expressions provide a flexible and intuitive way to convey fine-grained intent, enabling users to describe objects and their relationships in a manner that is more accessible than predefined labels.
Consequently, referring image segmentation has gained increasing attention in recent years, driven by its broad applicability in human-computer interaction~\cite{wang2019reinforced}, autonomous driving~\cite{codevilla2019exploring}, and smart education~\cite{xiao2025eduvqa}.

\begin{figure}[t]
  \centering
  \includegraphics[width=0.47\textwidth]{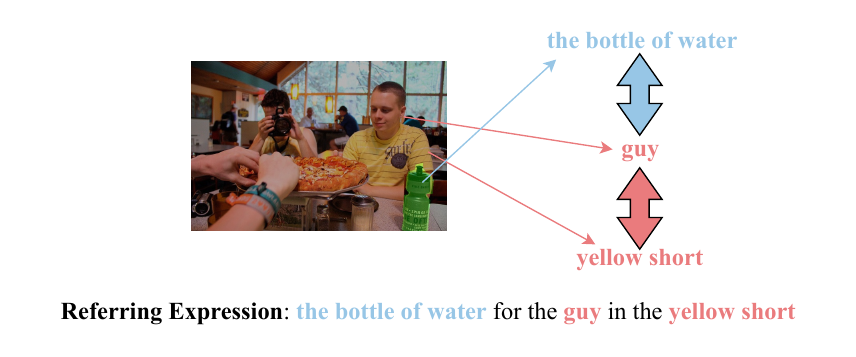}
  \caption{An example of referring image segmentation, where a free-form referring expression contains detailed attributes and complex relationships among multiple entities (e.g., \emph{bottle of water}, \emph{guy}, and \emph{yellow short}), requiring relational reasoning to accurately localize the target region.}
  \label{fig:intro}
\end{figure}

A key challenge in referring image segmentation lies in handling free-form expressions that contain detailed attributes and complex relationships~\cite{LI2026112549} among multiple entities, which complicates cross-modal reasoning.
As illustrated in Fig.~\ref{fig:intro}, a referring expression such as ``the bottle of water for the guy in the yellow short'' requires identifying multiple objects along with their associated attributes and reasoning about their relationships to accurately localize the corresponding region.
This observation raises a fundamental question: \emph{\textbf{how to effectively bridge free-form referring expressions with object-level visual representations through cross-modal reasoning.}}

\begin{figure}[t]
  \centering
  \includegraphics[width=0.49\textwidth]{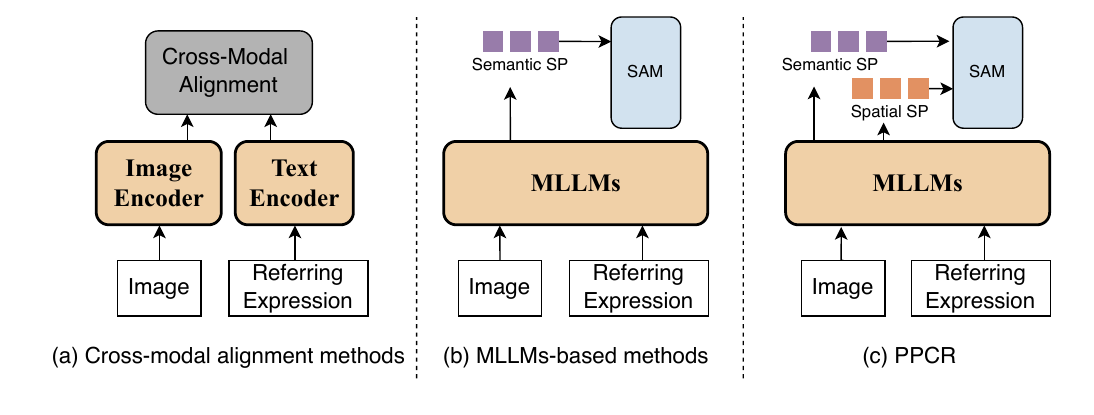}
  \caption{Illustration of different referring image segmentation paradigms. 
(a) Cross-modal alignment methods perform matching by directly aligning image and text representations.
(b) MLLMs-based methods generate Semantic Segmentation Prompt to guide segmentation, but lack explicit spatial grounding. 
(c) Our PPCR progressively generates Semantic and Spatial Segmentation Prompt, explicitly bridging semantic understanding and spatial grounding for accurate segmentation.}
  \label{fig:method}
\end{figure}

To address this problem, some methods~\cite{wang2022cris, yang2022lavt, liu2023caris, yue2024adaptive} mainly design various attention modules to fuse visual and linguistic features for cross-modal alignment, as shown in Fig.~\ref{fig:method} (a).
Although these approaches have achieved notable progress, their ability to explicitly reason from free-form referring expressions to object-level visual representations remains limited.
On the one hand, most of them focus on feature-level alignment, which makes it difficult to capture fine-grained relational information in free-form referring expressions.
On the other hand, these methods primarily rely on semantic similarity, limiting their capability to perceive implicit spatial relationships and interaction cues, and thus constraining localization performance in complex scenarios.

More recently, several studies~\cite{lai2024lisa, rasheed2024glamm} have explored MLLMs to enhance cross-modal reasoning, as shown in Fig.~\ref{fig:method} (b).
These approaches leverage MLLMs to interpret visual and textual inputs and produce \textbf{Semantic Segmentation Prompt (Semantic SP)} that guide downstream modules~\cite{kirillov2023segment} for localization and segmentation, such as LISA~\cite{lai2024lisa} and GLaMM~\cite{rasheed2024glamm}.
This paradigm represents a shift from cross-modal alignment toward explicit language-guided cross-modal reasoning.
Despite their strengths in commonsense reasoning and multimodal understanding, existing MLLMs-based approaches still exhibit limited spatial grounding capability.
Specifically, Semantic Segmentation Prompt typically convey only high-level semantic information, without an explicit reasoning process that links language understanding to object-level spatial localization.
As a result, such Semantic Segmentation Prompt remain weakly associated with spatial constraints, emphasizing semantic relevance while failing to encode explicit cues about target location, spatial extent, or relational structure.
Consequently, these methods often struggle in complex referring scenarios involving multiple similar objects and fine-grained attribute or relational descriptions, as well as producing accurate and complete segmentation masks, making it difficult to reliably bridge free-form referring expressions with object-level visual representations.

In this Paper, we propose PPCR, a \textbf{P}rogressive \textbf{P}rompt-guided \textbf{C}ross-modal \textbf{R}easoning framework for Referring Image Segmentation, as shown in Fig.~\ref{fig:method} (c).
PPCR constructs a \textbf{Semantic Understan-ding-Spatial Grounding-Instance Segmentation} reasoning pipe-line that progressively bridges free-form referring expressions with object-level visual representations, effectively mimicking human cognitive processes by decomposing the task into \emph{what}, \emph{where}, and \emph{how to segment}.
Specifically, PPCR begins with \textbf{Semantic Understanding} by employing MLLMs to generate Semantic Segmentation Prompt that capture key semantic cues of the target object described by the referring expression.
Based on this semantic context, PPCR performs \textbf{Spatial Grounding} by further generating Spatial Segmentation Prompt to reason about object location and spatial extent, enabling a progressive transition from semantic understanding to spatial grounding.
Finally, PPCR achieves \textbf{Instance Segmentation} by jointly integrating Semantic and Spatial Segmentation Prompt into the segmentation module (e.g., SAM~\cite{kirillov2023segment}) for accurate target localization and segmentation.
By explicitly structuring the reasoning process in a progressive manner, PPCR enables more reliable object-level grounding and thus improves referring image segmentation in complex scenarios.

The main contributions of this work are summarized as follows:
\begin{itemize}

  \item We propose PPCR, a \textbf{P}rogressive \textbf{P}rompt-guided \textbf{C}ross-modal \textbf{R}easoning framework for referring image segmentation, which explicitly bridges free-form referring expressions with object-level visual representations.

  \item We introduce a progressive prompt-guided reasoning scheme that jointly integrates Semantic and Spatial Segmentation Prompts into the segmentation module for accurate target localization and segmentation.

  \item Extensive experiments demonstrate that PPCR outperforms state-of-the-art methods, particularly in scenarios with multiple similar objects and fine-grained attribute or relational descriptions, and produces more accurate and complete segmentation masks.

\end{itemize}

\section{Related Work}
\label{sec:relatedwork}
\subsection{Referring Image Segmentation}
Referring image segmentation (RIS)~\cite{ding2025multimodal} aims to localize and segment the target object in an image given a natural language expression, requiring joint modeling of language semantics and object-level visual representations.
Early RIS methods~\cite{hu2016segmentation,liu2017recurrent,li2018referring, jiao2021two} relied on CNN-LSTM architectures to encode visual and textual features and fused them via recurrent interaction or iterative refinement.
Subsequent works~\cite{ye2019cross,yang2022lavt,zhang2022coupalign} design Transformer-based architectures with specialized attention modules to strengthen cross-modal alignment, enabling finer word-pixel interactions.
Another line of research~\cite{ding2021vision,wu2022language,tang2023contrastive, yuan2025text} introduces Learnable Query to drive Transformer Decoders for cross-modal interaction.
Despite their progress, these methods largely remain at feature-level alignment and provide limited explicit reasoning over relational and spatial structures in referring expressions.
With the success of large-scale Vision-Language Models~\cite{zhang2024vision}, recent approaches~\cite{wang2022cris,li2024mutually,wang2024cm} leverage CLIP~\cite{radford2021learning} pre-training to enhance cross-modal alignment in RIS.

More recently, multimodal large language models (MLLMs) have been explored for reasoning-based segmentation by generating language-guided prompts~\cite{lai2024lisa,wang2025segllm,rasheed2024glamm,zhang2024next, xia2024gsva}.
While these methods improve semantic understanding, they often rely on Semantic
Segmentation Prompt and do not explicitly connect language understanding with precise spatial grounding.
In contrast, our work progressively structures cross-modal reasoning to integrate semantic understanding, spatial grounding, and instance segmentation for RIS.

\begin{figure*}[t]
  \centering
  \includegraphics[width=0.75\textwidth]{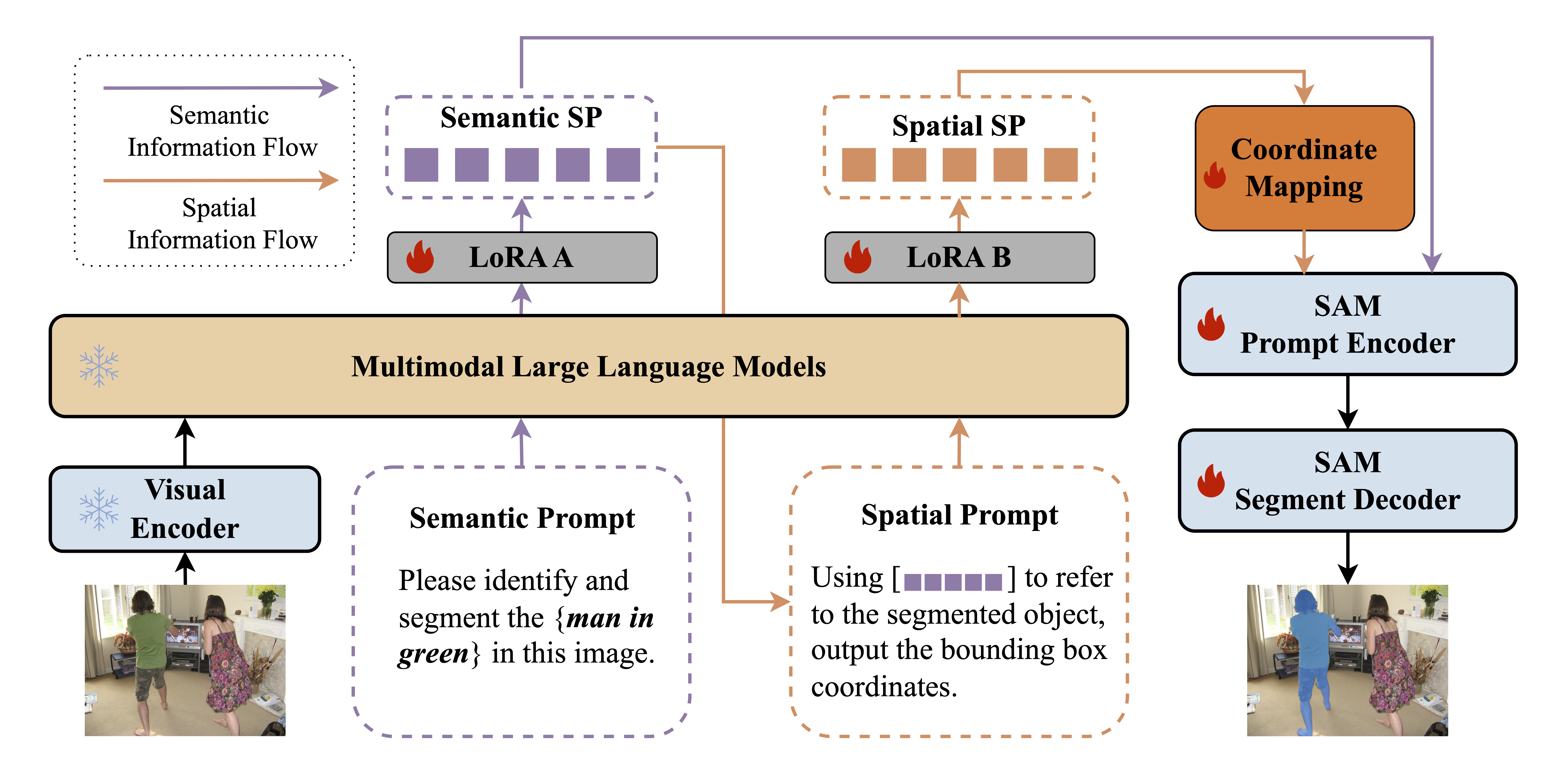}
  \caption{Overview of PPCR for referring image segmentation. 
The framework follows a \textbf{Semantic Understanding-Spatial Grounding-Instance Segmentation} pipeline. 
Given an image and a referring expression, a multimodal large language model first generates a Semantic Segmentation Prompt (Semantic SP) via the LoRA A branch, which then guides the generation of a Spatial Segmentation Prompt (Spatial SP) via the LoRA B branch. 
The Spatial Segmentation Prompt is mapped to bounding box coordinates, and is jointly used with the Semantic Segmentation Prompt by SAM for instance segmentation. 
Purple and orange arrows denote semantic and spatial information flows.}
\label{fig:framework}
\end{figure*}

\subsection{Prompt Learning for Image Segmentation}

Image segmentation~\cite{Long_2015_CVPR, Noh_2015_ICCV, He_2019_ICCV, He_2017_ICCV} is a fundamental task in computer vision, aiming to assign pixel-level labels to image regions for fine-grained understanding. 
Recent studies explore prompt learning as an effective paradigm to adapt pre-trained models for segmentation, which can be broadly categorized into text-based and visual-based prompting.

Text-based prompting formulates segmentation as a language-guided task by introducing textual descriptions as prompts. 
For example, PODA~\cite{fahes2023poda} employs manually designed prompts to describe target domains, enabling zero-shot domain adaptation by aligning generated features with domain-specific characteristics.

Visual-based prompting incorporates learnable prompt tokens or input transformations to guide segmentation models. 
Representative approaches, such as VPT~\cite{10.1007/978-3-031-19827-4_41} and EXPRES~\cite{Das_2023_CVPR}, introduce prompt tokens into vision Transformers for downstream segmentation tasks. 
Subsequent work~\cite{Liu_2023_CVPR} designs prompt modules to enhance feature representations, while other methods~\cite{NEURIPS2022_9f09f316, NEURIPS2023_157c30da} leverage context-based prompts or task-specific perturbations for improved adaptability.

The emergence of promptable segmentation models, such as SAM~\cite{kirillov2023segment}, further advances this paradigm by enabling segmentation conditioned on points, boxes, masks, and text. 
Follow-up works, including FastSAM~\cite{zhao2023fastsegment}, MobileSAM~\cite{zhang2023fastersegmentanythinglightweight}, and EfficientSAM~\cite{Xiong_2024_CVPR}, improve efficiency through lightweight designs, while HQ-SAM~\cite{NEURIPS2023_5f828e38} and PA-SAM~\cite{10687602} enhance mask quality via prompt refinement. 
Other extensions, such as SegGPT~\cite{wang2023seggptsegmentingcontext}, Grounded-SAM~\cite{ren2024groundedsamassemblingopenworld}, Prompt-RIS~\cite{10657777}, and SEEM~\cite{NEURIPS2023_3ef61f7e}, further extend prompt modalities and improve flexibility.

Despite these advances, existing prompt learning methods mainly treat prompts as direct guidance for segmentation, without explicitly structuring the reasoning process from semantic understanding to spatial localization. 
In contrast, our work organizes prompt generation into a progressive reasoning pipeline, bridging \emph{what}, \emph{where}, and \emph{how to segment}.

\section{Method}
\label{sec:method}
\subsection{Overview}
We propose PPCR, a \textbf{P}rogressive \textbf{P}rompt-guided \textbf{C}ross-modal \textbf{R}eas\-oning framework for referring image segmentation.
As shown in Fig.~\ref{fig:framework}, PPCR explicitly structures the reasoning process into a \textbf{Semantic Understanding-Spatial Grounding-Instance Segmentation} pipeline, progressively bridging free-form referring expressions with object-level visual representations.

Given an input image $I$ and a referring expression $T$, PPCR performs cross-modal reasoning through three stages. 
First, \textbf{Semantic Understanding} generates a Semantic Segmentation Prompt that captures key semantic cues of the target object (i.e., \emph{what} to segment). 
Second, \textbf{Spatial Grounding} produces a Spatial Segmentation Prompt conditioned on the semantic prompt to infer object location and spatial extent (i.e., \emph{where} to segment). 
Finally, \textbf{Instance Segmentation} integrates both Semantic and Spatial Segmentation Prompts into a segmentation model to predict the target mask.

\subsection{Semantic Understanding}

\begin{figure}[t]
  \centering
  \includegraphics[width=0.47\textwidth]{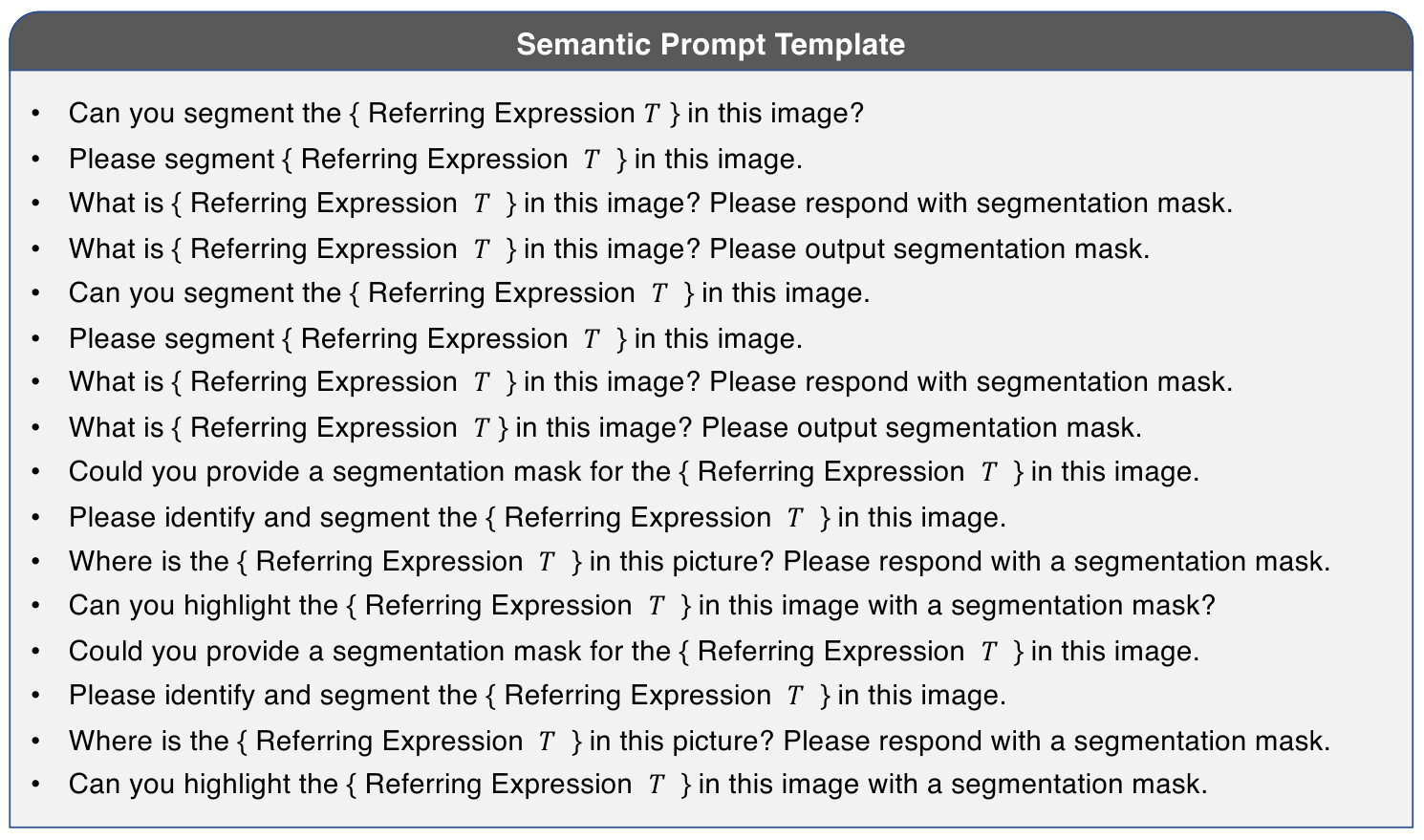}
  \caption{Illustration of the Semantic Prompt Template used in the Semantic Understanding stage.}
  \label{fig:T1}
\end{figure}

The semantic understanding stage aims to determine \emph{what} the referring expression describes, including object category, attributes, and relational cues.

Given an input image $I \in \mathbb{R}^{3 \times H \times W}$, a frozen visual encoder extracts visual features:
\begin{equation}
\mathbf{F}_v = F_{\mathrm{VE}}(I).
\end{equation}

The referring expression $T$ is combined with an semantic prompt template to form a semantic prompt $P_{\text{txt}}$, which provides structured guidance for the multimodal large language model.
As illustrated in Fig.~\ref{fig:T1}, we adopt a small set of structured templates to guide the multimodal large language model to generate segmentation-aware descriptions.
These templates are collected from existing MLLMs-based RIS works (e.g., LISA~\cite{lai2024lisa} and GLaMM~\cite{rasheed2024glamm}) and further extended to improve robustness across diverse expressions. During training, a template is randomly sampled for each referring expression to enhance generalization.

The multimodal large language model $F_{\mathrm{LLM}}$ takes visual features $\mathbf{F}_v$ and the semantic prompt $P_{\text{txt}}$ as input and generates a Semantic Segmentation Prompt:
\begin{equation}
P_{\text{txt}}^{\text{seg}} = F_{\mathrm{LLM}}(\mathbf{F}_v, P_{\text{txt}}).
\end{equation}

The generated Semantic Segmentation Prompt encodes high-level semantic information of the target, including object identity, attributes, and relevant contextual relations. 
It serves as a compact representation of \emph{what to segment}, but does not explicitly specify spatial location or extent.

This stage provides semantic grounding for subsequent reasoning and acts as the foundation for spatial localization in the next stage.

\subsection{Spatial Grounding}

\begin{figure}[h]
  \centering
  \includegraphics[width=0.47\textwidth]{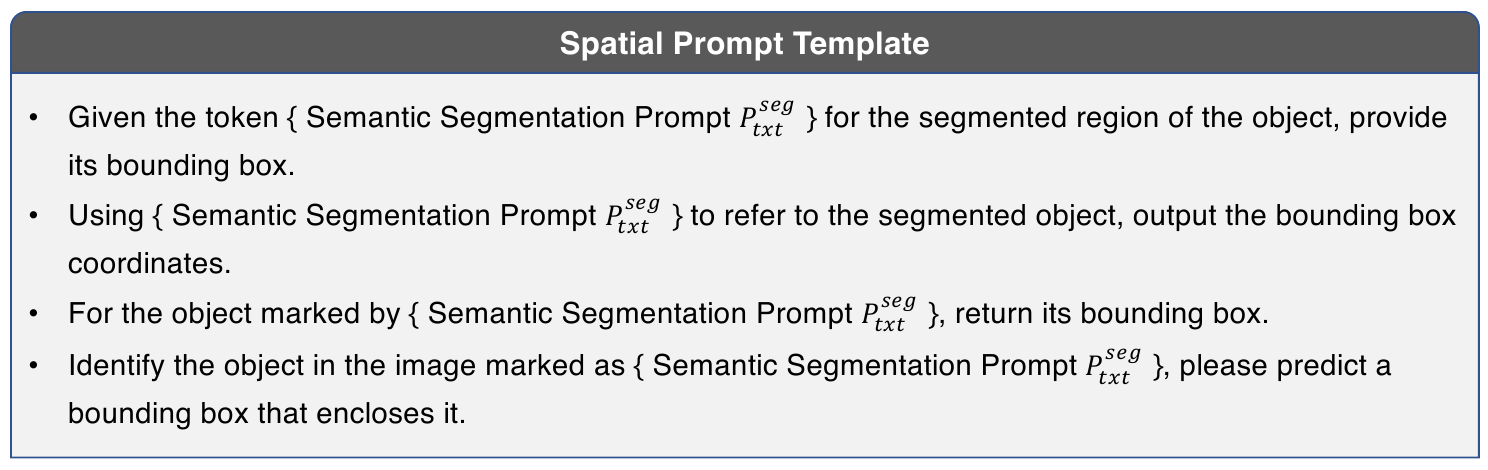}
  \caption{Illustration of the Spatial Prompt Template used in the Spatial Grounding stage.}
  \label{fig:T2}
\end{figure}

The spatial grounding stage aims to determine \emph{where} the target object is located, by explicitly reasoning about its spatial position and extent.

To achieve this, we construct a spatial prompt $P_{\text{box}}$ conditioned on the Semantic Segmentation Prompt $P_{\text{txt}}^{\text{seg}}$, enabling a progressive transition from semantic understanding to spatial localization. 
Specifically, as illustrated in Fig.~\ref{fig:T2}, we design a small set of simple spatial prompt templates that take $P_{\text{txt}}^{\text{seg}}$ as input and require the multimodal large language model to predict the corresponding bounding box. 
Unlike the semantic prompt templates in the previous stage, which are collected from existing works, these templates are manually designed in our method, as the goal here is to convert the semantic prompt into an explicit spatial localization request. 
During training, a template is randomly sampled to improve robustness to different instruction forms.

The multimodal large language model generates a Spatial Segmentation Prompt as:
\begin{equation}
P_{\text{box}}^{\text{seg}} = F_{\mathrm{LLM}}(\mathbf{F}_v, P_{\text{box}}).
\end{equation}

The Spatial Segmentation Prompt focuses on the position and extent of the target object in the image. 
By conditioning on $P_{\text{txt}}^{\text{seg}}$, this stage performs a progressive semantic-to-spatial transition, rather than predicting spatial information independently.

This stage establishes an explicit bridge between semantic understanding and spatial localization, enabling the model to progressively refine the target representation from \emph{what} to \emph{where}.

\subsection{Instance Segmentation}

In the final stage, PPCR performs instance segmentation by jointly integrating the Semantic Segmentation Prompt and the Spatial Segmentation Prompt, so as to generate the final pixel-level mask of the target object.

Since the Spatial Segmentation Prompt $P_{\text{box}}^{\text{seg}}$ produced in the previous stage is still represented as a high-dimensional embedding, it cannot be directly used as a box prompt for segmentation. 
To address this issue, we introduce a lightweight mapping module $F_{\mathrm{Map}}$ to transform $P_{\text{box}}^{\text{seg}}$ into normalized bounding box coordinates:
\begin{equation}
\tilde{\mathbf{b}} = F_{\mathrm{Map}}(P_{\text{box}}^{\text{seg}}), \quad \tilde{\mathbf{b}} \in [0,1]^4,
\end{equation}
where $\tilde{\mathbf{b}}=(\tilde{x}_1,\tilde{y}_1,\tilde{x}_2,\tilde{y}_2)$ denotes the normalized coordinates of the top-left and bottom-right corners of the predicted bounding box. 
Here, each coordinate is constrained to the range $[0,1]$, making the prediction independent of the input image resolution.
In practice, $F_{\mathrm{Map}}$ is implemented as a lightweight MLP to regress the bounding box coordinates from $P_{\text{box}}^{\text{seg}}$.

The normalized box coordinates are then rescaled to the original image size to obtain pixel-space coordinates:
\begin{equation}
\hat{\mathbf{b}} = (\tilde{x}_1 W,\tilde{y}_1 H,\tilde{x}_2 W,\tilde{y}_2 H),
\end{equation}
where $W$ and $H$ denote the width and height of the input image, respectively. 
In this way, $\hat{\mathbf{b}}$ can serve as the Spatial Segmentation Prompt for the downstream segmentation model.

Given the Semantic Segmentation Prompt $P_{\text{txt}}^{\text{seg}}$ and the box coordinates $\hat{\mathbf{b}}$, we then employ a segmentation model, such as SAM~\cite{kirillov2023segment}, to predict the final mask. 
Specifically, a frozen SAM image encoder first extracts multi-scale visual features $\mathbf{F}_v^{\mathrm{SAM}}$ from the input image. 
Meanwhile, the SAM prompt encoder takes $P_{\text{txt}}^{\text{seg}}$ and $\hat{\mathbf{b}}$ as prompt inputs, encoding the semantic cue and spatial cue into prompt features for mask prediction. 
Based on the encoded prompts and image features, the SAM mask decoder outputs the final segmentation result:
\begin{equation}
\hat{M} = F_{\mathrm{MDec}}^{\mathrm{SAM}}\!\left(
F_{\mathrm{PEnc}}^{\mathrm{SAM}}(P_{\text{txt}}^{\text{seg}}, \hat{\mathbf{b}}),
\mathbf{F}_v^{\mathrm{SAM}}
\right),
\end{equation}
where $F_{\mathrm{PEnc}}^{\mathrm{SAM}}(\cdot)$ and $F_{\mathrm{MDec}}^{\mathrm{SAM}}(\cdot)$ denote the prompt encoder and mask decoder of SAM, respectively, and $\hat{M}$ is the predicted segmentation mask.

By jointly introducing Semantic and Spatial Segmentation Prompts into the segmentation model, this stage completes the progressive transition from \emph{what} and \emph{where} to \emph{how to segment}, enabling more accurate and controllable target segmentation.

\subsection{Training Strategy and Optimization}

Although PPCR performs progressive reasoning during inference, an effective training process requires jointly optimizing semantic understanding, spatial grounding, and instance segmentation. 
To this end, we adopt a parameter-efficient fine-tuning strategy together with a multi-task training objective.

\textbf{Parameter-Efficient Prompt Generation.}
To adapt the multimodal large language model for Semantic and Spatial Segmentation Prompts generation while preserving its general multimodal capabilities, we employ Low-Rank Adaptation (LoRA)~\cite{hu2022lora}. 
Specifically, we introduce two task-specific LoRA branches: one for Semantic Segmentation Prompt generation (LoRA A), and the other for Spatial Segmentation Prompt generation (LoRA B). 
The main parameters of the MLLMs are kept frozen, and only the low-rank adaptation parameters are updated during training, enabling efficient and stable task adaptation. 
In this way, the model can learn task-specific prompt generation for both \emph{what} and \emph{where} without affecting its general reasoning ability.

\textbf{Joint Training Objective.}
The model is trained with a joint objective that supervises both instance segmentation and spatial localization. 
For instance segmentation, we adopt a binary cross-entropy loss $\mathcal{L}_{\mathrm{BCE}}$ and a Dice loss $\mathcal{L}_{\mathrm{Dice}}$ to ensure accurate pixel-level mask prediction. 
For spatial grounding, the predicted bounding box $\tilde{\mathbf{b}}$ is supervised using an $\ell_1$ regression loss $\mathcal{L}_{\ell_1}$, together with a classification loss $\mathcal{L}_{\mathrm{CE}}$ to encourage accurate localization.

The overall training objective is defined as:
\begin{equation}
\mathcal{L}_{\mathrm{total}} =
\lambda_1 \mathcal{L}_{\mathrm{BCE}} +
\lambda_2 \mathcal{L}_{\mathrm{Dice}} +
\lambda_3 \mathcal{L}_{\ell_1} +
\lambda_4 \mathcal{L}_{\mathrm{CE}},
\end{equation}
where $\lambda_1, \lambda_2, \lambda_3,$ and $\lambda_4$ are weighting coefficients that balance different loss terms.

This joint optimization strategy enables coordinated learning across semantic understanding, spatial grounding, and instance segmentation, corresponding to the progressive reasoning process from \emph{what} to \emph{where} and \emph{how to segment}.

\section{Experiment}
\label{sec:exp}

\subsection{Datastes and Metrics}
\textbf{Datasets.}
We evaluate PPCR on three widely used benchmark datasets for referring image segmentation: RefCOCO~\cite{refcoco}, RefCOCO+~\cite{refcoco}, and RefCOCOg~\cite{refcocog1,refcocog2}.
All three datasets are built on the MS COCO dataset and provide natural language expressions referring to specific target objects in images.

RefCOCO contains 19,994 images with 142,210 referring expressions, where most expressions describe the target object using simple attributes and spatial cues.

RefCOCO+ includes 19,992 images and 141,564 referring expressions, and removes explicit location words, thus emphasizing fine-grained visual attributes of the referent object.

RefCOCOg comprises 26,711 images with 104,560 referring expressions, featuring longer and more complex descriptions that require understanding relationships among multiple objects.

\textbf{Metrics.}
Following common practice in referring image segmentation, we adopt overall Intersection over Union (oIoU) and mean Intersection over Union (mIoU) to evaluate model performance.
oIoU computes the ratio between the total intersection area and the total union area over all samples, reflecting the overall segmentation quality while being sensitive to large objects.
In contrast, mIoU averages the IoU scores across all referring expressions, providing a more balanced evaluation for individual instances.

\subsection{Implementation Details}
PPCR is built on LLaVA-v1.5-7B with a CLIP-pretrained ViT-L/14 visual encoder.
For instance segmentation, we adopt SAM with a ViT-H/14 backbone, where the prompt encoder and mask decoder are initialized from the original SAM weights.
A sigmoid activation is used in the coordinate mapping layer to constrain predicted bounding boxes to $[0,1]$.

Parameter-Efficient fine-tuning is employed to adapt the MLLMs for prompt generation.
The semantic and spatial LoRA branches are initialized with GLaMM fine-tuned weights and GranD pre-trained weights, respectively.
The model is trained for 5 epochs using AdamW with a learning rate of $3\times10^{-4}$ and a WarmupDecayLR schedule.
Training is implemented in PyTorch with DeepSpeed (ZeRO-2).
We use a batch size of 8 and four loss terms with weights $\lambda_1=2$, $\lambda_2=0.5$, $\lambda_3=2$, and $\lambda_4=0.5$. 
The adopted lossesare commonly used for segmentation and box regression, and the weights are introduced to balance segmentation and localization objectives.
All experiments are conducted on four NVIDIA L20 GPUs.

\subsection{Comparison with State-of-the-art Methods}

\begin{table*}[t]
\centering
\caption{Comparison of referring image segmentation methods on RefCOCO, RefCOCO+, and RefCOCOg. \dag means the results are excerpted from the original paper.}
\label{tab:ris_comparison_all}
\setlength{\tabcolsep}{2pt}
\renewcommand{\arraystretch}{1.1}
\resizebox{\textwidth}{!}{
\begin{tabular}{c c cccccc cccccc cccc}
\toprule
\multirow{3}{*}{Method} & \multirow{3}{*}{MLLMs}
& \multicolumn{6}{c}{RefCOCO}
& \multicolumn{6}{c}{RefCOCO+}
& \multicolumn{4}{c}{RefCOCOg} \\
\cmidrule(lr){3-8}\cmidrule(lr){9-14}\cmidrule(lr){15-18}

& 
& \multicolumn{3}{c}{oIoU} & \multicolumn{3}{c}{mIoU}
& \multicolumn{3}{c}{oIoU} & \multicolumn{3}{c}{mIoU}
& \multicolumn{2}{c}{oIoU} & \multicolumn{2}{c}{mIoU} \\
\cmidrule(lr){3-5}\cmidrule(lr){6-8}
\cmidrule(lr){9-11}\cmidrule(lr){12-14}
\cmidrule(lr){15-16}\cmidrule(lr){17-18}

& 
& Val & TestA & TestB & Val & TestA & TestB
& Val & TestA & TestB & Val & TestA & TestB
& Val(U) & Test(U) & Val(U) & Test(U) \\
\midrule

CRIS$^{\dag}$~\cite{wang2022cris}      & --       
& -- & -- & -- & 70.47 & 73.18 & 66.10
& -- & -- & -- & 62.27 & 68.08 & 53.68
& -- & -- & 59.69 & 60.01 \\

LAVT~\cite{yang2022lavt}      & --       
& 72.73 & 75.82 & 68.79 & 74.76 & 76.89 & 70.94
& 62.14 & 68.38 & 55.10 & 65.81 & 70.97 & 59.23
& 61.23 & 62.09 & 63.34 & 63.63 \\

CoupAlign~\cite{zhang2022coupalign} & --       
& 74.70 & 77.76 & 70.58 & 75.49 & 77.15 & 63.98
& 62.92 & 68.34 & 56.69 & 66.17 & 71.56 & 60.23
& 62.84 & 62.22 & 63.76 & 64.32 \\

CARIS$^{\dag}$~\cite{liu2023caris} & --       
& 76.63 & 79.40 & 73.52 & -- & -- & --
& 68.03 & 73.70 & 60.41 & -- & -- & --
& 67.95 & 69.75 & -- & -- \\

ASDA$^{\dag}$~\cite{yue2024adaptive} & --       
& 75.06 & 77.14 & 71.36 & -- & -- & --
& 66.84 & 71.13 & 57.83 & -- & -- & --
& 65.73 & 66.45 & -- & -- \\

CMIRNet$^{\dag}$~\cite{xu2024cmirnet} & --       
& 78.24 & 80.44 & 75.22 & 78.98 & 80.73 & 76.56
& 69.82 & 75.33 & 62.07 & 72.50 & 77.00 & 66.86
& 70.36 & 72.09 & 72.60 & 73.18 \\

LISA~\cite{lai2024lisa}     & LLaVA-7B 
& 67.24 & 70.37 & 63.18 & 67.84 & 70.46 & 63.96
& 53.95 & 59.43 & 47.16 & 53.28 & 59.63 & 46.45
& 60.91 & 61.78 & 61.08 & 61.63 \\

NExT-Chat$^{\dag}$~\cite{zhang2024next}     & 7B 
& 74.7 & 78.9 & 69.5 & -- & -- & --
& 65.1 & 71.9 & 56.7 & -- & -- & --
& 67.0 & 67.0 & -- & -- \\

GSVA$^{\dag}$~\cite{xia2024gsva}     & LLaVA-7B 
& 77.2 & 78.9 & 73.5 & -- & -- & --
& 65.9 & 69.6 & 59.8 & -- & -- & --
& 72.7 & 73.3 & -- & -- \\

SAM4MLLMs-PQPP$^{\dag}$~\cite{chen2024sam4mllm}     & Qwen-VL-7B
& 77.1 & 80.9 & 72.5 & -- & -- & --
& 71.5 & 76.8 & 64.7 & -- & -- & --
& 74.5 & 75.2 & -- & -- \\

SegLLM~\cite{wang2025segllm}   & LLaVA-7B 
& 79.32 & 81.69 & 76.58 & 79.55 & 82.70 & 78.12
& 70.78 & 73.41 & 62.11 & 73.70 & 77.22 & 67.43
& 72.30 & 73.24 & 74.19 & 74.43 \\

GLaMM~\cite{rasheed2024glamm}    & LLaVA-7B 
& 79.50 & 81.92 & 76.90 & 80.76 & 82.78 & \textbf{78.36}
& 72.10 & 76.60 & 64.60 & 73.76 & 78.05 & 68.38
& 73.66 & 74.70 & 75.11 & 75.66 \\

\textbf{PPCR}       & LLaVA-7B 
& \textbf{80.10} & \textbf{83.55} & \textbf{77.60}
& \textbf{81.10} & \textbf{83.69} & 78.33
& \textbf{72.50} & \textbf{78.52} & \textbf{65.55}
& \textbf{74.34} & \textbf{79.56} & \textbf{69.33}
& \textbf{74.92} & \textbf{75.70} & \textbf{75.94} & \textbf{76.10} \\

Absolute& 
& +0.60 & +1.63 & +0.70 & +0.34 & +0.91 & -0.03
& +0.40 & +1.92 & +0.95 & +0.58 & +1.51 & +0.95
& +1.26 & +1.00 & +0.84 & +0.44 \\
\bottomrule
\end{tabular}
}
\end{table*}

Tab.~\ref{tab:ris_comparison_all} reports quantitative comparisons between PPCR and representative referring image segmentation methods on RefCOCO, RefCOCO+, and RefCOCOg using both oIoU and mIoU.
For a fair comparison, we adopt the same LLaVA-v1.5-7B backbone when comparing with MLLMs-based methods.

\textbf{Results on RefCOCO.}
On RefCOCO, PPCR achieves overall better performance than GLaMM across the three splits.
Specifically, PPCR improves oIoU by +0.60\%, +1.63\%, and +0.70\% on Val, TestA, and TestB, respectively.
For mIoU, PPCR achieves gains of +0.34\% on Val and +0.91\% on TestA, while showing a slight decrease of 0.03\% on TestB.
These results suggest that the proposed progressive reasoning strategy improves both overall localization accuracy and instance-level segmentation quality in most standard referring scenarios, while maintaining competitive performance on the more challenging split.

\textbf{Results on RefCOCO+.}
RefCOCO+ focuses on fine-grained attribute descriptions and introduces higher semantic ambiguity.
Compared with GLaMM, PPCR improves oIoU by +0.40\%, +1.92\%, and +0.95\% on Val, TestA, and TestB, respectively.
Correspondingly, PPCR improves mIoU by +0.58\%, +1.51\%, and +0.95\% on the three splits.
The consistent gains across both metrics suggest that progressively generating Semantic and Spatial Segmentation Prompt effectively strengthens attribute-aware reasoning and spatial discrimination.

\textbf{Results on RefCOCOg.}
RefCOCOg contains longer referring expressions with complex relational structures.
PPCR achieves consistent improvements over GLaMM on both splits, with oIoU gains of +1.26\% on Val(U) and +1.00\% on Test(U).
For mIoU, PPCR improves performance by +0.84\% and +0.44\% on Val(U) and Test(U), respectively.
These results demonstrate that explicitly structuring the reasoning process is particularly beneficial for handling long expressions and multi-entity relational reasoning.

Overall, compared with cross-modal alignment methods  such as CRIS~\cite{wang2022cris}, LAVT~\cite{yang2022lavt}, CoupAlign~\cite{zhang2022coupalign}, CARIS~\cite{liu2023caris}, ASDA~\cite{yue2024adaptive} and CMIRNet~\cite{xu2024cmirnet}, as well as recent MLLMs-based methods including LISA~\cite{lai2024lisa}, NExT-Chat~\cite{zhang2024next}, GSVA~\cite{xia2024gsva}, SAM4MLLMs-PQPP~\cite{chen2024sam4mllm}, SegLLM~\cite{wang2025segllm} and GLaMM~\cite{rasheed2024glamm}, PPCR achieves the best overall performance.
More importantly, these results validate the effectiveness of the proposed progressive reasoning framework.
Instead of relying solely on implicit feature interaction or coarse prompt guidance, PPCR decomposes referring image segmentation into three successive stages: \textbf{Semantic Understanding-Spatial Grounding-Prompt-Guided Instance Segmentation}.
Such a design enables the model to first determine \emph{what} the target is, then infer \emph{where} it is located, and finally decide \emph{how to segment} it at the pixel level.
This progressive reasoning process is better aligned with the intrinsic requirements of RIS, and thus provides a more reliable solution for challenging cases involving ambiguous expressions, multiple similar objects, and complex spatial relations.

\subsection{Ablation Study}

\begin{table*}[h]
\centering
\caption{Ablation study on different reasoning mechanisms on RefCOCO, RefCOCO+, and RefCOCOg.}
\label{tab:reasoning_mechanisms}
\small
\setlength{\tabcolsep}{2pt}
\renewcommand{\arraystretch}{1.1}
\resizebox{\textwidth}{!}{
\begin{tabular}{c 
    ccc ccc 
    ccc ccc 
    cc cc}
\toprule
\multirow{3}{*}{Variant}
& \multicolumn{6}{c}{RefCOCO}
& \multicolumn{6}{c}{RefCOCO+}
& \multicolumn{4}{c}{RefCOCOg} \\
\cmidrule(lr){2-7}\cmidrule(lr){8-13}\cmidrule(lr){14-17}
& \multicolumn{3}{c}{oIoU} & \multicolumn{3}{c}{mIoU}
& \multicolumn{3}{c}{oIoU} & \multicolumn{3}{c}{mIoU}
& \multicolumn{2}{c}{oIoU} & \multicolumn{2}{c}{mIoU} \\
\cmidrule(lr){2-4}\cmidrule(lr){5-7}
\cmidrule(lr){8-10}\cmidrule(lr){11-13}
\cmidrule(lr){14-15}\cmidrule(lr){16-17}
& Val & TestA & TestB & Val & TestA & TestB
& Val & TestA & TestB & Val & TestA & TestB
& Val(U) & Test(U) & Val(U) & Test(U) \\
\midrule
w/o Spatial SP
& 79.50 & 81.92 & 76.90 & 80.76 & 82.78 & \textbf{78.36}
& 72.10 & 76.60 & 64.60 & 73.76 & 78.05 & 68.38
& 73.66 & 74.70 & 75.11 & 75.66 \\
w/ Spatial SP (RE)
& \textbf{80.22} & 82.76 & 76.66 & 80.99 & 82.98 & 77.96
& 70.93 & 78.04 & 64.25 & 73.35 & 78.81 & 67.45
& 74.30 & 75.08 & 75.17 & 75.73 \\
w/ Spatial SP (RE + $P_{\text{txt}}^{\text{seg}}$)
& 79.80 & 82.32 & 77.08 & 80.54 & 82.59 & 78.34
& 72.15 & 77.50 & \textbf{66.08} & 73.72 & 78.44 & 69.15
& 74.63 & 75.10 & 75.30 & 75.74 \\
PPCR
& 80.10 & \textbf{83.55} & \textbf{77.60} & \textbf{81.10} & \textbf{83.69} & 78.33
& \textbf{72.50} & \textbf{78.52} & 65.55 & \textbf{74.34} & \textbf{79.56} & \textbf{69.33}
& \textbf{74.92} & \textbf{75.70} & \textbf{75.94} & \textbf{76.10} \\
\bottomrule
\end{tabular}
}
\end{table*}

\textbf{Ablation Study on Reasoning Mechanisms.}
We analyze the effect of different reasoning designs by progressively introducing Semantic and Spatial Segmentation Prompts.
Results on RefCOCO, RefCOCO+, and RefCOCOg are summarized in Table~\ref{tab:reasoning_mechanisms}.
\textbf{w/o Spatial SP.}
This setting corresponds to the GLAMM~\cite{rasheed2024glamm}, where only the Semantic Segmentation Prompt is used.
Without explicit spatial cues, the model relies solely on semantic information, which limits its ability.
\textbf{w/ Spatial SP (RE).}
Spatial Segmentation Prompt is directly generated from the referring expression, without conditioning on the Semantic Segmentation Prompt.
Specifically, the model takes the raw expression and a predefined spatial prompt template (e.g., “\emph{Where is the target described by \{T\}? Please respond with a bounding box.}”) to predict the spatial location.
This design introduces explicit spatial cues and improves performance on RefCOCO.
However, since the spatial reasoning relies solely on the raw expression, the performance becomes unstable on RefCOCO+ and RefCOCOg, where expressions involve more complex attributes and relations.
This suggests that spatial prompts derived only from expressions may introduce ambiguity and unreliable localization.
\textbf{w/ Spatial SP (RE + $P_{\text{txt}}^{\text{seg}}$).}
Spatial Segmentation Prompt is generated by jointly conditioning on both the referring expression and the Semantic Segmentation Prompt $P_{\text{txt}}^{\text{seg}}$.
Specifically, the Semantic Segmentation Prompt is used to indicate the target region, and combined with the original expression to construct a spatial prompt template (e.g., “\emph{Take the region marked by \{$P_{\text{txt}}^{\text{seg}}$\} for \{T\}, and provide its bounding box.}”).
Compared with the previous variant, this design introduces semantic guidance into spatial reasoning, leading to more consistent improvements across datasets.
This demonstrates that incorporating semantic cues helps reduce ambiguity and provides a more reliable basis for spatial localization.
\textbf{PPCR.}
The full model further enforces a progressive reasoning process, where Spatial Segmentation Prompt are generated solely based on Semantic Segmentation Prompt.
This design achieves the best performance across most metrics and improves robustness under complex referring scenarios.
These results verify that explicitly structuring the reasoning process is critical for reliable referring image segmentation.

\begin{table*}[t]
\centering
\caption{Ablation study on Spatial Grounding on RefCOCO, RefCOCO+, and RefCOCOg.}
\label{tab:spatial_prompts}
\small
\setlength{\tabcolsep}{2pt}
\renewcommand{\arraystretch}{1.1}
\resizebox{\textwidth}{!}{
\begin{tabular}{c 
    ccc ccc 
    ccc ccc 
    cc cc}
\toprule
\multirow{3}{*}{Variant}
& \multicolumn{6}{c}{RefCOCO}
& \multicolumn{6}{c}{RefCOCO+}
& \multicolumn{4}{c}{RefCOCOg} \\
\cmidrule(lr){2-7}\cmidrule(lr){8-13}\cmidrule(lr){14-17}
& \multicolumn{3}{c}{oIoU} & \multicolumn{3}{c}{mIoU}
& \multicolumn{3}{c}{oIoU} & \multicolumn{3}{c}{mIoU}
& \multicolumn{2}{c}{oIoU} & \multicolumn{2}{c}{mIoU} \\
\cmidrule(lr){2-4}\cmidrule(lr){5-7}
\cmidrule(lr){8-10}\cmidrule(lr){11-13}
\cmidrule(lr){14-15}\cmidrule(lr){16-17}
& Val & TestA & TestB & Val & TestA & TestB
& Val & TestA & TestB & Val & TestA & TestB
& Val(U) & Test(U) & Val(U) & Test(U) \\
\midrule
w/ Attn-based Box Prediction
& 80.11 & 82.31 & 76.81 & 81.09 & 82.98 & 77.92
& 71.87 & 76.34 & 65.20 & 73.93 & 78.33 & 68.44
& 75.37 & 75.03 & 75.97 & 75.90 \\
w/ GT Box (Oracle)
& \textbf{88.32} & \textbf{88.93} & \textbf{86.84} & \textbf{87.79} & \textbf{88.12} & \textbf{86.72}
& \textbf{87.94} & \textbf{88.76} & \textbf{87.12} & \textbf{87.58} & \textbf{87.95} & \textbf{86.82}
& \textbf{87.45} & \textbf{87.58} & \textbf{85.58} & \textbf{85.73} \\
PPCR
& 80.10 & 83.55 & 77.60 & 81.10 & 83.69 & 78.33
& 72.50 & 78.52 & 65.55 & 74.34 & 79.56 & 69.33
& 74.92 & 75.70 & 75.94 & 76.10 \\
\bottomrule
\end{tabular}
}
\end{table*}

\textbf{Ablation Study on Spatial Grounding.}
We further investigate the impact of different Spatial Grounding strategies on referring image segmentation performance, focusing on how Spatial Segmentation Prompt are transformed into bounding box coordinates.
Results on RefCOCO, RefCOCO+, and RefCOCOg are reported in Table~\ref{tab:spatial_prompts}.
\textbf{w/ Attn-based Box Prediction.}
We replace the lightweight mapping module with a stack of self-attention and cross-attention layers to regress bounding box coordinates from visual features, where Spatial Segmentation Prompt serve as queries.
This design achieves comparable performance on some splits, indicating that attention-based reasoning can model spatial localization.
However, it introduces substantially higher model complexity and computational cost.
\textbf{w/ GT Box (Oracle).}
In this setting, ground-truth bounding box coordinates are directly used as spatial prompts together with Semantic Segmentation Prompt.
This oracle configuration achieves the best performance across all datasets and metrics, providing an upper bound for Spatial Segmentation Prompt.
The significant performance gap between this setting and ours Spatial Segmentation Prompt highlights both the importance of accurate spatial guidance and the remaining limitations of ours Spatial Segmentation Prompt.
\textbf{PPCR.}
The proposed method generates Spatial Segmentation Prompt conditioned on Semantic Segmentation Prompt and maps Spatial Segmentation Prompt to bounding box coordinates via a lightweight MLP-based mapping module.
Compared with attention-based box prediction, this design achieves slightly better overall performance while maintaining much lower computational cost.
These results indicate that explicitly structuring spatial grounding as a prompt generation and lightweight mapping process is both effective and efficient.
Overall, PPCR provides a favorable trade-off between accuracy and efficiency, demonstrating that lightweight and structured spatial grounding is more suitable for scalable referring image segmentation.

\subsection{Qualitative Results}

\begin{figure*}[t]
  \centering
  \includegraphics[width=0.8\textwidth]{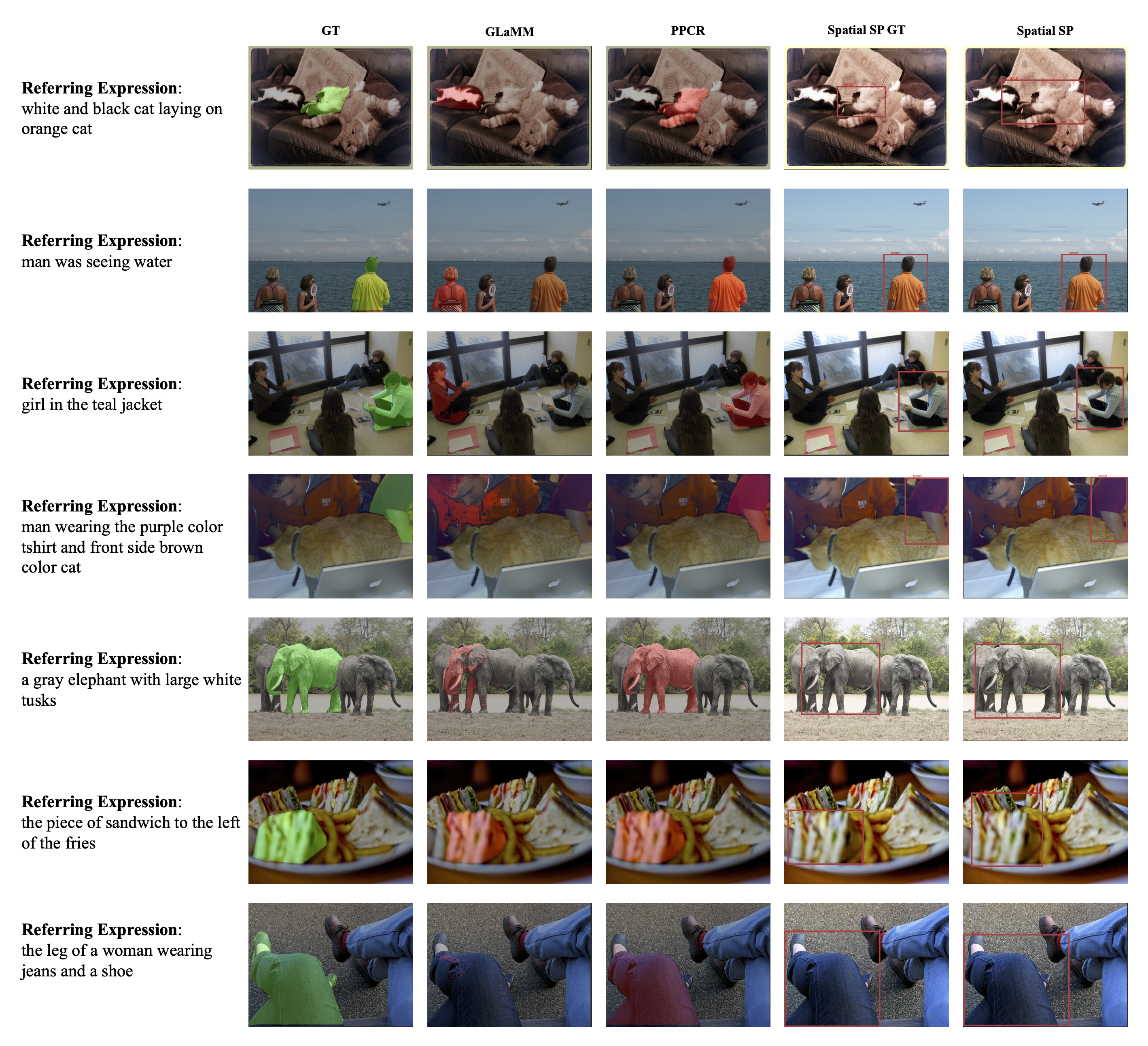}
  \caption{Qualitative comparisons on RefCOCOg. From left to right: Ground-Truth masks (GT), GLaMM results, PPCR results, the Spatial Segmentation Prompt Ground Truth (Spatial SP GT), and the Spatial Segmentation Prompt (Spatial SP).}
  \label{fig:qualitative}
\end{figure*}

To further validate the effectiveness of the proposed progressive prompt-guided cross-modal reasoning framework, we conduct qualitative comparisons on representative samples from RefCOCOg. 
Fig.~\ref{fig:qualitative} shows the Ground-Truth masks, the results of GLaMM~\cite{rasheed2024glamm}, the results of PPCR, the Ground-Truth Spatial Segmentation Prompt, and the predicted Spatial Segmentation Prompt.

As shown in Fig.~\ref{fig:qualitative}, PPCR consistently produces more accurate localization and more complete segmentation results than GLaMM in complex referring scenarios involving multiple objects and relational descriptions. 
Specifically, PPCR achieves more precise target localization (e.g., rows 1--4), benefiting from the progressive prompt-guided cross-modal reasoning framework. 
Moreover, PPCR generates more complete object masks (e.g., rows 5--7), while GLaMM tends to produce partial or coarse segmentation results. 
These observations indicate that progressively integrating semantic and spatial cues leads to better alignment between language and visual regions, thereby improving both localization accuracy and mask completeness.

\subsection{Failure Cases}

\begin{figure}[htp]
  \centering
  \includegraphics[width=0.47\textwidth]{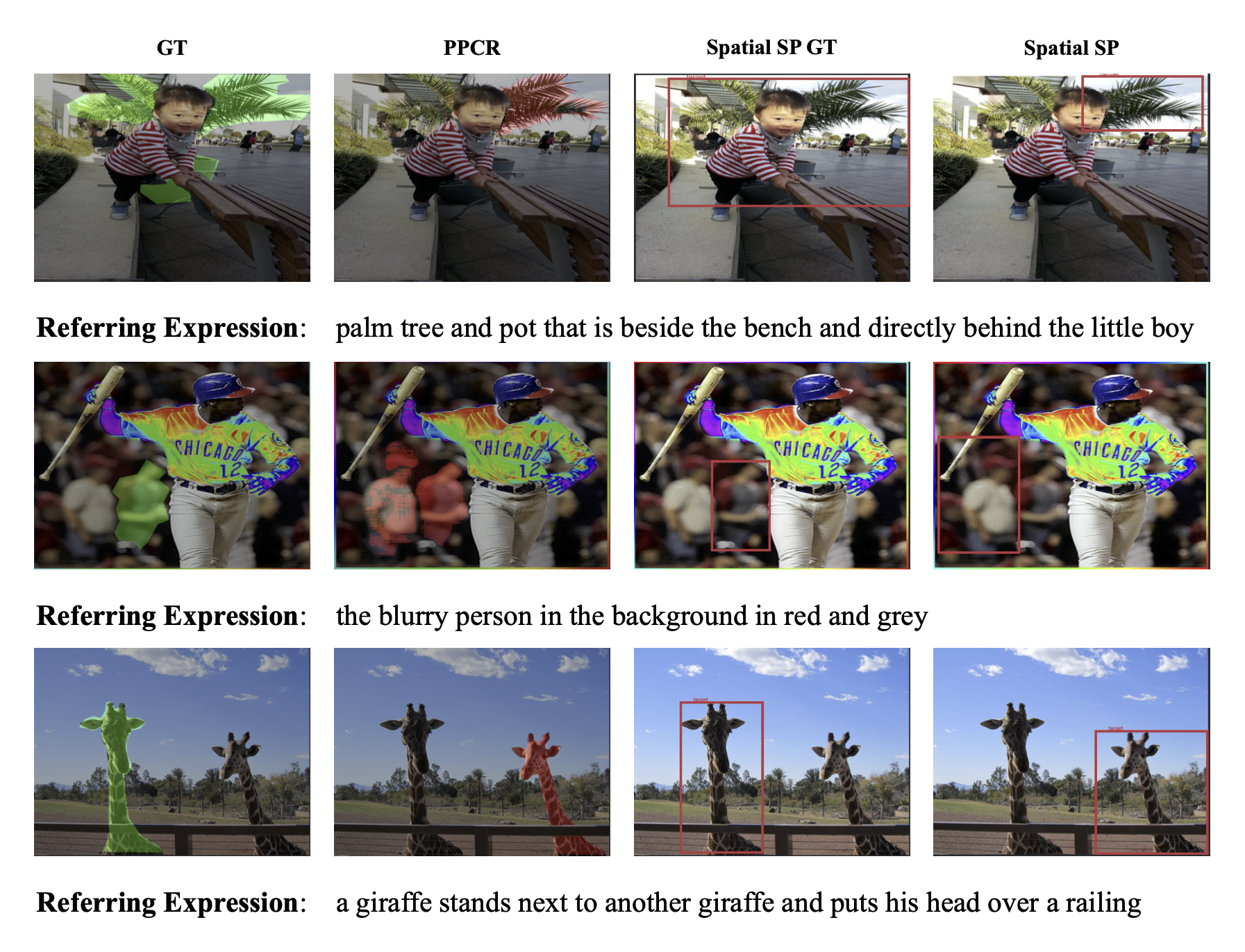}
  \caption{Failure cases of PPCR.}
  \label{fig:fc}
\end{figure}

We further analyze representative failure cases of PPCR in Fig.~\ref{fig:fc}. 
Failures mainly occur under three challenging conditions. 
When the target is partially occluded, incomplete visual evidence leads to inaccurate spatial grounding and fragmented masks. 
In scenes with blurred or low-quality backgrounds, weak visual discriminability hinders reliable matching between language and visual regions. 
Moreover, ambiguous or underspecified referring expressions prevent the model from uniquely identifying the target, resulting in incorrect localization. 
These results suggest that robust spatial grounding still depends on complete visual cues and unambiguous language descriptions.

\section{Conclusion}
In this paper, we propose PPCR, a progressive prompt-guided cross-modal reasoning framework for referring image segmentation. 
PPCR structures the reasoning process into a Semantic Understanding-Spatial Grounding-Instance Segmentation pipeline to bridge free-form referring expressions with object-level visual representations. 
Specifically, Semantic Segmentation Prompt are first generated to capture key semantic cues, and are then used to guide the generation of Spatial Segmentation Prompt for explicit localization. 
The Semantic and Spatial Segmentation Prompts are jointly integrated into the segmentation model to produce the final mask. 
Extensive experiments on RefCOCO, RefCOCO+, and RefCOCOg demonstrate the effectiveness and robustness of the proposed method. 
These results verify that progressively structuring semantic and spatial reasoning leads to more reliable referring image segmentation under complex scenarios.


\bibliographystyle{unsrt}  
\bibliography{references}

@article{Xiao_2025,
   title={Towards Visual Grounding: A Survey},
   ISSN={1939-3539},
   url={http://dx.doi.org/10.1109/TPAMI.2025.3630635},
   DOI={10.1109/tpami.2025.3630635},
   journal={IEEE Transactions on Pattern Analysis and Machine Intelligence},
   publisher={Institute of Electrical and Electronics Engineers (IEEE)},
   author={Xiao, Linhui and Yang, Xiaoshan and Lan, Xiangyuan and Wang, Yaowei and Xu, Changsheng},
   year={2025},
   pages={1–20} }

@inproceedings{wang2019reinforced,
  title={Reinforced cross-modal matching and self-supervised imitation learning for vision-language navigation},
  author={Wang, Xin and Huang, Qiuyuan and Celikyilmaz, Asli and Gao, Jianfeng and Shen, Dinghan and Wang, Yuan-Fang and Wang, William Yang and Zhang, Lei},
  booktitle={Proceedings of the IEEE/CVF conference on computer vision and pattern recognition},
  pages={6629--6638},
  year={2019}
}

@inproceedings{codevilla2019exploring,
  title={Exploring the limitations of behavior cloning for autonomous driving},
  author={Codevilla, Felipe and Santana, Eder and L{\'o}pez, Antonio M and Gaidon, Adrien},
  booktitle={Proceedings of the IEEE/CVF international conference on computer vision},
  pages={9329--9338},
  year={2019}
}

@article{xiao2025eduvqa,
  title={EduVQA: A multimodal Visual Question Answering framework for smart education},
  author={Xiao, Jiongen and Zhang, Zifeng},
  journal={Alexandria Engineering Journal},
  volume={122},
  pages={615--624},
  year={2025},
  publisher={Elsevier}
}

@ARTICLE{10613466,
  author={Li, Xiangtai and Ding, Henghui and Yuan, Haobo and Zhang, Wenwei and Pang, Jiangmiao and Cheng, Guangliang and Chen, Kai and Liu, Ziwei and Loy, Chen Change},
  journal={IEEE Transactions on Pattern Analysis and Machine Intelligence}, 
  title={Transformer-Based Visual Segmentation: A Survey}, 
  year={2024},
  volume={46},
  number={12},
  pages={10138-10163},
  doi={10.1109/TPAMI.2024.3434373}}

@article{LI2026112549,
title = {LGD: Leveraging generative descriptions for zero-shot referring image segmentation},
journal = {Pattern Recognition},
volume = {172},
pages = {112549},
year = {2026},
issn = {0031-3203},
doi = {https://doi.org/10.1016/j.patcog.2025.112549},
url = {https://www.sciencedirect.com/science/article/pii/S0031320325012129},
author = {Jiachen Li and Qing Xie and Renshu Gu and Jinyu Xu and Yongjian Liu and Xiaohan Yu}
}

@inproceedings{wang2022cris,
  title={Cris: Clip-driven referring image segmentation},
  author={Wang, Zhaoqing and Lu, Yu and Li, Qiang and Tao, Xunqiang and Guo, Yandong and Gong, Mingming and Liu, Tongliang},
  booktitle={Proceedings of the IEEE/CVF conference on computer vision and pattern recognition},
  pages={11686--11695},
  year={2022}
}

@inproceedings{yang2022lavt,
  title={Lavt: Language-aware vision transformer for referring image segmentation},
  author={Yang, Zhao and Wang, Jiaqi and Tang, Yansong and Chen, Kai and Zhao, Hengshuang and Torr, Philip HS},
  booktitle={Proceedings of the IEEE/CVF conference on computer vision and pattern recognition},
  pages={18155--18165},
  year={2022}
}

@inproceedings{liu2023caris,
  title={CARIS: Context-aware referring image segmentation},
  author={Liu, Sun-Ao and Zhang, Yiheng and Qiu, Zhaofan and Xie, Hongtao and Zhang, Yongdong and Yao, Ting},
  booktitle={Proceedings of the 31st ACM International Conference on Multimedia},
  pages={779--788},
  year={2023}
}

@inproceedings{lai2024lisa,
  title={Lisa: Reasoning segmentation via large language model},
  author={Lai, Xin and Tian, Zhuotao and Chen, Yukang and Li, Yanwei and Yuan, Yuhui and Liu, Shu and Jia, Jiaya},
  booktitle={Proceedings of the IEEE/CVF Conference on Computer Vision and Pattern Recognition},
  pages={9579--9589},
  year={2024}
}

@inproceedings{rasheed2024glamm,
  title={Glamm: Pixel grounding large multimodal model},
  author={Rasheed, Hanoona and Maaz, Muhammad and Shaji, Sahal and Shaker, Abdelrahman and Khan, Salman and Cholakkal, Hisham and Anwer, Rao M and Xing, Eric and Yang, Ming-Hsuan and Khan, Fahad S},
  booktitle={Proceedings of the IEEE/CVF Conference on Computer Vision and Pattern Recognition},
  pages={13009--13018},
  year={2024}
}

@inproceedings{kirillov2023segment,
  title={Segment anything},
  author={Kirillov, Alexander and Mintun, Eric and Ravi, Nikhila and Mao, Hanzi and Rolland, Chloe and Gustafson, Laura and Xiao, Tete and Whitehead, Spencer and Berg, Alexander C and Lo, Wan-Yen and others},
  booktitle={Proceedings of the IEEE/CVF international conference on computer vision},
  pages={4015--4026},
  year={2023}
}

@inproceedings{refcoco,
  title={Modeling context between objects for referring expression understanding},
  author={Nagaraja, Varun K and Morariu, Vlad I and Davis, Larry S},
  booktitle={Computer Vision--ECCV 2016: 14th European Conference, Amsterdam, The Netherlands, October 11--14, 2016, Proceedings, Part IV 14},
  pages={792--807},
  year={2016},
  organization={Springer}
}

@inproceedings{refcocog1,
  title={Referitgame: Referring to objects in photographs of natural scenes},
  author={Kazemzadeh, Sahar and Ordonez, Vicente and Matten, Mark and Berg, Tamara},
  booktitle={Proceedings of the 2014 conference on empirical methods in natural language processing (EMNLP)},
  pages={787--798},
  year={2014}
}

@inproceedings{refcocog2,
  title={Generation and comprehension of unambiguous object descriptions},
  author={Mao, Junhua and Huang, Jonathan and Toshev, Alexander and Camburu, Oana and Yuille, Alan L and Murphy, Kevin},
  booktitle={Proceedings of the IEEE conference on computer vision and pattern recognition},
  pages={11--20},
  year={2016}
}

@article{ding2025multimodal,
  title={Multimodal referring segmentation: A survey},
  author={Ding, Henghui and Tang, Song and He, Shuting and Liu, Chang and Wu, Zuxuan and Jiang, Yu-Gang},
  journal={arXiv preprint arXiv:2508.00265},
  year={2025}
}

@inproceedings{hu2016segmentation,
  title={Segmentation from natural language expressions},
  author={Hu, Ronghang and Rohrbach, Marcus and Darrell, Trevor},
  booktitle={European conference on computer vision},
  pages={108--124},
  year={2016},
  organization={Springer}
}

@inproceedings{liu2017recurrent,
  title={Recurrent multimodal interaction for referring image segmentation},
  author={Liu, Chenxi and Lin, Zhe and Shen, Xiaohui and Yang, Jimei and Lu, Xin and Yuille, Alan},
  booktitle={Proceedings of the IEEE international conference on computer vision},
  pages={1271--1280},
  year={2017}
}

@inproceedings{li2018referring,
  title={Referring image segmentation via recurrent refinement networks},
  author={Li, Ruiyu and Li, Kaican and Kuo, Yi-Chun and Shu, Michelle and Qi, Xiaojuan and Shen, Xiaoyong and Jia, Jiaya},
  booktitle={Proceedings of the IEEE Conference on Computer Vision and Pattern Recognition},
  pages={5745--5753},
  year={2018}
}

@inproceedings{ye2019cross,
  title={Cross-modal self-attention network for referring image segmentation},
  author={Ye, Linwei and Rochan, Mrigank and Liu, Zhi and Wang, Yang},
  booktitle={Proceedings of the IEEE/CVF conference on computer vision and pattern recognition},
  pages={10502--10511},
  year={2019}
}

@article{zhang2022coupalign,
  title={Coupalign: Coupling word-pixel with sentence-mask alignments for referring image segmentation},
  author={Zhang, Zicheng and Zhu, Yi and Liu, Jianzhuang and Liang, Xiaodan and Ke, Wei},
  journal={Advances in Neural Information Processing Systems},
  volume={35},
  pages={14729--14742},
  year={2022}
}

@inproceedings{ding2021vision,
  title={Vision-language transformer and query generation for referring segmentation},
  author={Ding, Henghui and Liu, Chang and Wang, Suchen and Jiang, Xudong},
  booktitle={Proceedings of the IEEE/CVF international conference on computer vision},
  pages={16321--16330},
  year={2021}
}

@inproceedings{wu2022language,
  title={Language as queries for referring video object segmentation},
  author={Wu, Jiannan and Jiang, Yi and Sun, Peize and Yuan, Zehuan and Luo, Ping},
  booktitle={Proceedings of the IEEE/CVF Conference on Computer Vision and Pattern Recognition},
  pages={4974--4984},
  year={2022}
}

@inproceedings{tang2023contrastive,
  title={Contrastive grouping with transformer for referring image segmentation},
  author={Tang, Jiajin and Zheng, Ge and Shi, Cheng and Yang, Sibei},
  booktitle={Proceedings of the IEEE/CVF conference on computer vision and pattern recognition},
  pages={23570--23580},
  year={2023}
}

@article{li2024mutually,
  title={Mutually-guided hierarchical multi-modal feature learning for referring image segmentation},
  author={Li, Jiachen and Xie, Qing and Chang, Xiaojun and Xu, Jinyu and Liu, Yongjian},
  journal={ACM Transactions on Multimedia Computing, Communications and Applications},
  volume={20},
  number={12},
  pages={1--18},
  year={2024},
  publisher={ACM New York, NY}
}

@article{wang2024cm,
  title={Cm-masksd: Cross-modality masked self-distillation for referring image segmentation},
  author={Wang, Wenxuan and He, Xingjian and Zhang, Yisi and Guo, Longteng and Shen, Jiachen and Li, Jiangyun and Liu, Jing},
  journal={IEEE Transactions on Multimedia},
  volume={26},
  pages={6906--6916},
  year={2024},
  publisher={IEEE}
}

@inproceedings{wang2025segllm,
  title={Segllm: Multi-round reasoning segmentation with large language models},
  author={Wang, XuDong and Zhang, Shaolun and Li, Shufan and Li, Kehan and Kallidromitis, Konstantinos and Kato, Yusuke and Kozuka, Kazuki and Darrell, Trevor},
  booktitle={The Thirteenth International Conference on Learning Representations},
  year={2025}
}

@inproceedings{jiao2021two,
  title={Two-stage visual cues enhancement network for referring image segmentation},
  author={Jiao, Yang and Jie, Zequn and Luo, Weixin and Chen, Jingjing and Jiang, Yu-Gang and Wei, Xiaolin and Ma, Lin},
  booktitle={Proceedings of the 29th ACM international conference on multimedia},
  pages={1331--1340},
  year={2021}
}

@inproceedings{yue2024adaptive,
  title={Adaptive selection based referring image segmentation},
  author={Yue, Pengfei and Lin, Jianghang and Zhang, Shengchuan and Hu, Jie and Lu, Yilin and Niu, Hongwei and Ding, Haixin and Zhang, Yan and Jiang, Guannan and Cao, Liujuan and others},
  booktitle={Proceedings of the 32nd ACM International Conference on Multimedia},
  pages={1101--1110},
  year={2024}
}

@inproceedings{yuan2025text,
  title={Text-promptable propagation for referring medical image sequence segmentation},
  author={Yuan, Runtian and Chen, Mohan and Xu, Jilan and Zhou, Ling and Li, Qingqiu and Zhang, Yuejie and Feng, Rui and Zhang, Tao and Gao, Shang},
  booktitle={Proceedings of the 33rd ACM International Conference on Multimedia},
  pages={362--371},
  year={2025}
}

@article{xu2024cmirnet,
  title={CMIRNet: Cross-modal interactive reasoning network for referring image segmentation},
  author={Xu, Mingzhu and Xiao, Tianxiang and Liu, Yutong and Tang, Haoyu and Hu, Yupeng and Nie, Liqiang},
  journal={IEEE Transactions on Circuits and Systems for Video Technology},
  volume={35},
  number={4},
  pages={3234--3249},
  year={2024},
  publisher={IEEE}
}

@inproceedings{xia2024gsva,
  title={Gsva: Generalized segmentation via multimodal large language models},
  author={Xia, Zhuofan and Han, Dongchen and Han, Yizeng and Pan, Xuran and Song, Shiji and Huang, Gao},
  booktitle={Proceedings of the IEEE/CVF Conference on Computer Vision and Pattern Recognition},
  pages={3858--3869},
  year={2024}
}

@inproceedings{chen2024sam4mllm,
  title={Sam4mllm: Enhance multi-modal large language model for referring expression segmentation},
  author={Chen, Yi-Chia and Li, Wei-Hua and Sun, Cheng and Wang, Yu-Chiang Frank and Chen, Chu-Song},
  booktitle={European Conference on Computer Vision},
  pages={323--340},
  year={2024},
  organization={Springer}
}

@inproceedings{zhang2024next,
  title={NExT-Chat: an LMM for chat, detection and segmentation},
  author={Zhang, Ao and Yao, Yuan and Ji, Wei and Liu, Zhiyuan and Chua, Tat-Seng},
  booktitle={Proceedings of the 41st International Conference on Machine Learning},
  pages={60116--60133},
  year={2024}
}

@article{hu2022lora,
  title={Lora: Low-rank adaptation of large language models.},
  author={Hu, Edward J and Shen, Yelong and Wallis, Phillip and Allen-Zhu, Zeyuan and Li, Yuanzhi and Wang, Shean and Wang, Liang and Chen, Weizhu and others},
  journal={Iclr},
  volume={1},
  number={2},
  pages={3},
  year={2022}
}

@inproceedings{fahes2023poda,
  title={Poda: Prompt-driven zero-shot domain adaptation},
  author={Fahes, Mohammad and Vu, Tuan-Hung and Bursuc, Andrei and P{\'e}rez, Patrick and De Charette, Raoul},
  booktitle={Proceedings of the IEEE/CVF international conference on computer vision},
  pages={18623--18633},
  year={2023}
}

@InProceedings{10.1007/978-3-031-19827-4_41,
author="Jia, Menglin
and Tang, Luming
and Chen, Bor-Chun
and Cardie, Claire
and Belongie, Serge
and Hariharan, Bharath
and Lim, Ser-Nam",
editor="Avidan, Shai
and Brostow, Gabriel
and Ciss{\'e}, Moustapha
and Farinella, Giovanni Maria
and Hassner, Tal",
title="Visual Prompt Tuning",
booktitle="Computer Vision -- ECCV 2022",
year="2022",
publisher="Springer Nature Switzerland",
address="Cham",
pages="709--727",
isbn="978-3-031-19827-4"
}

@InProceedings{Das_2023_CVPR,
    author    = {Das, Rajshekhar and Dukler, Yonatan and Ravichandran, Avinash and Swaminathan, Ashwin},
    title     = {Learning Expressive Prompting With Residuals for Vision Transformers},
    booktitle = {Proceedings of the IEEE/CVF Conference on Computer Vision and Pattern Recognition (CVPR)},
    month     = {June},
    year      = {2023},
    pages     = {3366-3377}
}

@InProceedings{Liu_2023_CVPR,
    author    = {Liu, Weihuang and Shen, Xi and Pun, Chi-Man and Cun, Xiaodong},
    title     = {Explicit Visual Prompting for Low-Level Structure Segmentations},
    booktitle = {Proceedings of the IEEE/CVF Conference on Computer Vision and Pattern Recognition (CVPR)},
    month     = {June},
    year      = {2023},
    pages     = {19434-19445}
}

@inproceedings{NEURIPS2022_9f09f316,
 author = {Bar, Amir and Gandelsman, Yossi and Darrell, Trevor and Globerson, Amir and Efros, Alexei},
 booktitle = {Advances in Neural Information Processing Systems},
 editor = {S. Koyejo and S. Mohamed and A. Agarwal and D. Belgrave and K. Cho and A. Oh},
 pages = {25005--25017},
 publisher = {Curran Associates, Inc.},
 title = {Visual Prompting via Image Inpainting},
 url = {https://proceedings.neurips.cc/paper_files/paper/2022/file/9f09f316a3eaf59d9ced5ffaefe97e0f-Paper-Conference.pdf},
 volume = {35},
 year = {2022}
}

@inproceedings{NEURIPS2023_157c30da,
 author = {Ma, Xinhong and Wang, Yiming and Liu, Hao and Guo, Tianyu and Wang, Yunhe},
 booktitle = {Advances in Neural Information Processing Systems},
 editor = {A. Oh and T. Naumann and A. Globerson and K. Saenko and M. Hardt and S. Levine},
 pages = {6690--6702},
 publisher = {Curran Associates, Inc.},
 title = {When Visual Prompt Tuning Meets Source-Free Domain Adaptive Semantic Segmentation},
 url = {https://proceedings.neurips.cc/paper_files/paper/2023/file/157c30da6a988e1cbef2095f7b9521db-Paper-Conference.pdf},
 volume = {36},
 year = {2023}
}

@misc{zhao2023fastsegment,
      title={Fast Segment Anything}, 
      author={Xu Zhao and Wenchao Ding and Yongqi An and Yinglong Du and Tao Yu and Min Li and Ming Tang and Jinqiao Wang},
      year={2023},
      eprint={2306.12156},
      archivePrefix={arXiv},
      primaryClass={cs.CV},
      url={https://arxiv.org/abs/2306.12156}, 
}

@misc{zhang2023fastersegmentanythinglightweight,
      title={Faster Segment Anything: Towards Lightweight SAM for Mobile Applications}, 
      author={Chaoning Zhang and Dongshen Han and Yu Qiao and Jung Uk Kim and Sung-Ho Bae and Seungkyu Lee and Choong Seon Hong},
      year={2023},
      eprint={2306.14289},
      archivePrefix={arXiv},
      primaryClass={cs.CV},
      url={https://arxiv.org/abs/2306.14289}, 
}

@InProceedings{Xiong_2024_CVPR,
    author    = {Xiong, Yunyang and Varadarajan, Bala and Wu, Lemeng and Xiang, Xiaoyu and Xiao, Fanyi and Zhu, Chenchen and Dai, Xiaoliang and Wang, Dilin and Sun, Fei and Iandola, Forrest and Krishnamoorthi, Raghuraman and Chandra, Vikas},
    title     = {EfficientSAM: Leveraged Masked Image Pretraining for Efficient Segment Anything},
    booktitle = {Proceedings of the IEEE/CVF Conference on Computer Vision and Pattern Recognition (CVPR)},
    month     = {June},
    year      = {2024},
    pages     = {16111-16121}
}

@inproceedings{NEURIPS2023_5f828e38,
 author = {Ke, Lei and Ye, Mingqiao and Danelljan, Martin and liu, Yifan and Tai, Yu-Wing and Tang, Chi-Keung and Yu, Fisher},
 booktitle = {Advances in Neural Information Processing Systems},
 editor = {A. Oh and T. Naumann and A. Globerson and K. Saenko and M. Hardt and S. Levine},
 pages = {29914--29934},
 publisher = {Curran Associates, Inc.},
 title = {Segment Anything in High Quality},
 url = {https://proceedings.neurips.cc/paper_files/paper/2023/file/5f828e38160f31935cfe9f67503ad17c-Paper-Conference.pdf},
 volume = {36},
 year = {2023}
}

@INPROCEEDINGS{10687602,
  author={Xie, Zhaozhi and Guan, Bochen and Jiang, Weihao and Yi, Muyang and Ding, Yue and Lu, Hongtao and Zhang, Lei},
  booktitle={2024 IEEE International Conference on Multimedia and Expo (ICME)}, 
  title={PA-SAM: Prompt Adapter SAM for High-Quality Image Segmentation}, 
  year={2024},
  volume={},
  number={},
  pages={1-6},
  doi={10.1109/ICME57554.2024.10687602}}

@misc{wang2023seggptsegmentingcontext,
      title={SegGPT: Segmenting Everything In Context}, 
      author={Xinlong Wang and Xiaosong Zhang and Yue Cao and Wen Wang and Chunhua Shen and Tiejun Huang},
      year={2023},
      eprint={2304.03284},
      archivePrefix={arXiv},
      primaryClass={cs.CV},
      url={https://arxiv.org/abs/2304.03284}, 
}

@misc{ren2024groundedsamassemblingopenworld,
      title={Grounded SAM: Assembling Open-World Models for Diverse Visual Tasks}, 
      author={Tianhe Ren and Shilong Liu and Ailing Zeng and Jing Lin and Kunchang Li and He Cao and Jiayu Chen and Xinyu Huang and Yukang Chen and Feng Yan and Zhaoyang Zeng and Hao Zhang and Feng Li and Jie Yang and Hongyang Li and Qing Jiang and Lei Zhang},
      year={2024},
      eprint={2401.14159},
      archivePrefix={arXiv},
      primaryClass={cs.CV},
      url={https://arxiv.org/abs/2401.14159}, 
}

@inproceedings{NEURIPS2023_3ef61f7e,
 author = {Zou, Xueyan and Yang, Jianwei and Zhang, Hao and Li, Feng and Li, Linjie and Wang, Jianfeng and Wang, Lijuan and Gao, Jianfeng and Lee, Yong Jae},
 booktitle = {Advances in Neural Information Processing Systems},
 editor = {A. Oh and T. Naumann and A. Globerson and K. Saenko and M. Hardt and S. Levine},
 pages = {19769--19782},
 publisher = {Curran Associates, Inc.},
 title = {Segment Everything Everywhere All at Once},
 url = {https://proceedings.neurips.cc/paper_files/paper/2023/file/3ef61f7e4afacf9a2c5b71c726172b86-Paper-Conference.pdf},
 volume = {36},
 year = {2023}
}

@INPROCEEDINGS{10657777,
  author={Shang, Chao and Song, Zichen and Qiu, Heqian and Wang, Lanxiao and Meng, Fanman and Li, Hongliang},
  booktitle={2024 IEEE/CVF Conference on Computer Vision and Pattern Recognition (CVPR)}, 
  title={Prompt-Driven Referring Image Segmentation with Instance Contrasting}, 
  year={2024},
  volume={},
  number={},
  pages={4124-4134},
  doi={10.1109/CVPR52733.2024.00395}}

@InProceedings{Long_2015_CVPR,
author = {Long, Jonathan and Shelhamer, Evan and Darrell, Trevor},
title = {Fully Convolutional Networks for Semantic Segmentation},
booktitle = {Proceedings of the IEEE Conference on Computer Vision and Pattern Recognition (CVPR)},
month = {June},
year = {2015}
}

@InProceedings{Noh_2015_ICCV,
author = {Noh, Hyeonwoo and Hong, Seunghoon and Han, Bohyung},
title = {Learning Deconvolution Network for Semantic Segmentation},
booktitle = {Proceedings of the IEEE International Conference on Computer Vision (ICCV)},
month = {December},
year = {2015}
}

@InProceedings{He_2019_ICCV,
author = {He, Junjun and Deng, Zhongying and Qiao, Yu},
title = {Dynamic Multi-Scale Filters for Semantic Segmentation},
booktitle = {Proceedings of the IEEE/CVF International Conference on Computer Vision (ICCV)},
month = {October},
year = {2019}
}

@InProceedings{He_2017_ICCV,
author = {He, Kaiming and Gkioxari, Georgia and Dollar, Piotr and Girshick, Ross},
title = {Mask R-CNN},
booktitle = {Proceedings of the IEEE International Conference on Computer Vision (ICCV)},
month = {Oct},
year = {2017}
}

@inproceedings{radford2021learning,
  title={Learning transferable visual models from natural language supervision},
  author={Radford, Alec and Kim, Jong Wook and Hallacy, Chris and Ramesh, Aditya and Goh, Gabriel and Agarwal, Sandhini and Sastry, Girish and Askell, Amanda and Mishkin, Pamela and Clark, Jack and others},
  booktitle={International conference on machine learning},
  pages={8748--8763},
  year={2021},
  organization={PmLR}
}

@article{zhang2024vision,
  title={Vision-language models for vision tasks: A survey},
  author={Zhang, Jingyi and Huang, Jiaxing and Jin, Sheng and Lu, Shijian},
  journal={IEEE transactions on pattern analysis and machine intelligence},
  volume={46},
  number={8},
  pages={5625--5644},
  year={2024},
  publisher={IEEE}
}

\end{document}